\documentclass[letterpaper]{article} 
\usepackage[draft]{aaai25}  
\usepackage{times}  
\usepackage{helvet}  
\usepackage{courier}  
\usepackage[hyphens]{url}  
\usepackage{graphicx} 
\usepackage{natbib}  
\usepackage{caption} 
\usepackage{algorithm}

\usepackage{newfloat}
\usepackage{listings}
\DeclareCaptionStyle{ruled}{labelfont=normalfont,labelsep=colon,strut=off} 
\floatstyle{ruled}
\newfloat{listing}{tb}{lst}{}
\floatname{listing}{Listing}
\usepackage{hyperref}
\usepackage{appendix}
\usepackage{svg}
\usepackage{xcolor}
\usepackage{amsfonts}
\usepackage{multirow}
\usepackage{array}
\usepackage{amsmath}
\usepackage{booktabs}
\usepackage{algpseudocode}

\newcommand\norm[1]{\lVert#1\rVert^2_2}

\newcommand{\figref}[1]{Figure~\ref{#1}}

\newif\ifboldvector 
\boldvectortrue

\newcommand{\bvector}[1]{%
  \ifboldvector
    \mathbf{#1}     
  \else
    #1              
  \fi
}

\title{Zero-Shot Conditioning of Score-Based Diffusion Models by Neuro-Symbolic Constraints}
\author {
    Davide Scassola\textsuperscript{\rm 1,2},
    Sebastiano Saccani\textsuperscript{\rm 2},
    Ginevra Carbone\textsuperscript{\rm 2},
    Luca Bortolussi\textsuperscript{\rm 1}
}
\affiliations {
    \textsuperscript{\rm 1}AI\textsc{lab}, University of Trieste, Trieste, Italy\\
    \textsuperscript{\rm 2}Aindo, AREA Science Park, Trieste, Italy\\
    davide.scassola@phd.units.it, sebastiano@aindo.com, ginevracoal@gmail.com, lbortolussi@units.it
}

\begin{document}

\maketitle

\begin{abstract}
Score-based diffusion models have emerged as effective approaches for both conditional and unconditional generation. Still conditional generation is based on either a specific training of a conditional model or classifier guidance, which requires training  a noise-dependent classifier, even when a classifier for uncorrupted data is given.
We propose a method that, given a pre-trained unconditional score-based generative model, samples from the conditional distribution under arbitrary logical constraints, without requiring additional training. Differently from other zero-shot techniques, that rather aim at generating valid conditional samples, our method is designed for approximating the true conditional distribution.
Firstly, we show how to manipulate the learned score in order to sample from an un-normalized distribution conditional on a user-defined constraint. Then, we define a flexible and numerically stable neuro-symbolic framework for encoding soft logical constraints. Combining these two ingredients we obtain a general, but approximate, conditional sampling algorithm. We further developed effective heuristics aimed at improving the approximation.      
Finally, we show the effectiveness of our approach in approximating conditional distributions for various types of constraints and data: tabular data, images and time series.
\end{abstract}

%

\section{Introduction}

Score-based \cite{song2019generative} and diffusion \cite{ho2020denoising,Sohl-DicksteinW15} generative models based on deep neural networks have proven effective in modelling complex high-dimensional distributions in various domains. 
Controlling these models in order to obtain desirable features in samples is often required, still most conditional models require additional constraint-specific training in order to perform conditional sampling. 
This represents a limit since either one needs to train an extremely flexible conditional model (as those based on text prompts), or alternatively to train a conditional model for any specific constraint  to be enforced. Moreover, these conditional models often lack robustness, since the constraint is only learned through labelled data, even when the constraint is a user-defined function.
As a consequence, zero-shot conditional generation with arbitrary but formal logical constraints specified at inference time is currently hard. This would be useful in different contexts, for example: 
\begin{itemize}
    \item \textit{Tabular data}: generation of entries that obey  formal requirements described by logical formulas, without a specific training for every formula.
    \item \textit{Surrogate models}: using unconditional surrogate models to efficiently sample imposing physical constraints, or exploring scenarios defined by additional constraints.
\end{itemize}

Text prompt conditioned image generation with diffusion models \cite{rombach2022high} has proven extremely flexible and effective, still it lacks fine grained control and requires a massive amount of labelled samples.
Recent work focuses on methods to perform guided diffusion on images, without the need to retrain a noise-dependent classifier \cite{bansal2023universal,graikos2022diffusion,kadkhodaie2021stochastic, nair2023steered}, but the performance of the approximation of the conditional distribution is never evaluated.

In this article we develop a method for sampling from pre-trained unconditional score-based generative models, enforcing arbitrary user-defined logical constraints, that does not require additional training. Despite being originally designed for tabular data, we also show the application of our method to images and time series.
In summary, we present the following key contributions:

\begin{itemize}
    
    \item We develop a 
    zero-shot method for applying constraints to pre-trained unconditional score-based generative models. 
    The method enables sampling approximately from the conditional distribution given a soft constraint.
    \item We define a general neuro-symbolic language for building soft constraints that corresponds to logical formulas. These constraints are numerically stable, satisfy convenient logical properties, and can be relaxed/hardened arbitrarily through a control parameter.
    \item We test our method on different types of datasets and constraints, showing good performance on \textit{approximating conditional distributions} on tabular data, while previous methods were rather meant to obtain high-quality samples that satisfy some given constraints.
    \item Comparing our method with a state-of-the-art method \cite{bansal2023universal}, we gather evidence that plug-and-play conditioning techniques designed for images are not necessarily suited for modelling the true conditional distribution. Moreover, we show our neuro-symbolic language to be useful for defining constraints also within this other method.
    \item We show that our method allows to sample conditioning on  constraints that involve multiple data instances.
\end{itemize}

\section{Background}
\subsection{Score-based generative models}

Score-based generative models \cite{song2019generative,sde}, are a class of models developed in recent years, closely related to diffusion probabilistic models \cite{Sohl-DicksteinW15,ho2020denoising}.
Given the objective to sample from the distribution $p(\bvector{x})$ that generated the data, these models aim at estimating the (\textit{Stein}) score of $p(\bvector{x})$, defined as $\bvector{s}(\bvector{x}) := \nabla_\bvector{x} \ln{p(\bvector{x})}$ and then use sampling techniques that exploit the knowledge of the score of the distribution.
There are several methods for estimating the score: score matching \cite{hyvarinen2005estimation}, sliced score matching \cite{song2020sliced}, and denoising score matching \cite{vincent2011connection}. Denoising score matching is probably the most popular one, it uses corrupted data samples $\tilde{\bvector{x}}$ in order to estimate the score of the distribution for different levels of added noise, which is in practice necessary for sampling in high dimensional spaces \cite{song2019generative}.

Given a neural network $\bvector{s}_\theta(\bvector{x},t)$ and a diffusion process $q_t(\tilde{\bvector{x}} | \bvector{x})$ defined for $t \in [0,1]$ such that $q_0(\tilde{\bvector{x}} | \bvector{x}) \approx \delta(\bvector{x})$ (no corruption) and $q_1(\tilde{\bvector{x}} | \bvector{x})$ is a fixed prior distribution (e.g., a Gaussian), the denoising score matching loss is:
$$\mathbb{E}_{t\sim u(0,1), \bvector{x} \sim p(\bvector{x}), \tilde{\bvector{x}} \sim q_t(\tilde{\bvector{x}}|\bvector{x})} \norm{\bvector{s}_\theta(\tilde{\bvector{x}},t) - \nabla_{\tilde{\bvector{x}}} \ln{q_t(\tilde{\bvector{x}} | \bvector{x})}}$$

With sufficient data and model capacity, denoising score matching ensures $\bvector{s}_\theta(\bvector{x},t) \approx \nabla_{\bvector{x}} \ln{p_t(\bvector{x})}$ for almost all $\bvector{x}$ and $t$, where $p_t(\bvector{x}) := \int{q_t(\bvector{x} | \bvector{x}_0)p_0(\bvector{x}_0)d\bvector{x}_0}$ is the distribution of the data for different levels of added noise.
Given the estimate of the time/noise dependent score $\bvector{s}_\theta(\bvector{x},t)$, one can resort to different techniques for sampling from $p(\bvector{x}) = p_0(\bvector{x})$ as annealed Langevin dynamics \cite{song2019generative}, denoising diffusion probabilistic models \cite{ho2020denoising} or stochastic differential equations \cite{sde}, a generalization of the aforementioned approaches.



\subsection{Conditional sampling with score-based models}
Given a joint distribution $p(\bvector{x}, \bvector{y})$, one is often interested in sampling from the conditional distribution $p(\bvector{x} | \bvector{y})$, where $\bvector{y}$ is for example a label. While it is possible to directly model the conditional distribution (as usually done in many generative models), in this case by estimating $\nabla_{\bvector{x}} \ln{p_t(\bvector{x} | \bvector{y})}$, score-based generative models allow conditional sampling without explicit training of the conditional generative model. Applying the Bayesian rule $p_t(\bvector{x} | \bvector{y}) = \frac{p_t(\bvector{y} | \bvector{x})p_t(\bvector{x})}{p(\bvector{y})}$ one can observe that:
$$\nabla_{\bvector{x}} \ln{p_t(\bvector{x} | \bvector{y})} = \nabla_{\bvector{x}} \ln{p_t(\bvector{x})} + \nabla_{\bvector{x}} \ln{p_t(\bvector{y} | \bvector{x})}$$
It follows that one can obtain the conditional score from the unconditional score by separately training a noise-dependent classifier $p_t(\bvector{y} | \bvector{x})$. This technique is known as ``guidance", and it has been used for class conditional image generation \cite{dhariwal2021diffusion}.

\section{Related Work}

Controllable generation with score-based models is discussed in \citet{sde}, where they both treat the case when the noise conditional classifier $p_t(\bvector{y} | \tilde{\bvector{x}})$ can be trained and when it is not available. In that case, a method for obtaining such an estimate without the need of training auxiliary models is discussed and applied to conditional generation tasks such as inpainting and colorization. Still, the estimate is applicable only assuming 
the possibility to define $\bvector{y}_t$ such that $p(\bvector{y}_t|\bvector{y})$ and $p(\bvector{x}_t|\bvector{y}_t)$ are tractable.
Earlier works explored the combination of unconditional latent variables generative models such as Generative Adversarial Networks (GANs) \cite{goodfellow2014generative} or Variational Autoencoders (VAEs) \cite{kingma2013auto} with constraints to produce conditional samples \cite{engel2017latent}. As for DDPMs (denoising diffusion probabilistic models), recent research has focused on leveraging pre-trained unconditional models as priors for solving inverse problems as in \citet{kadkhodaie2021stochastic} where they use DDPMs for solving linear inverse imaging problems, and \citet{graikos2022diffusion}, \citet{bansal2023universal}, \citet{nair2023steered} where they generalize to a generic guidance. 
Despite these works focused on sampling high-quality samples that satisfy some given properties, it was not verified if samples followed the correct conditional distributions, a more difficult task that is relevant for example when dealing with tabular data and time series. Given that these methods are often based on the introduction of an optimization phase in the original sampling process, it is not guaranteed that the true conditional distribution will be well approximated. Moreover, these methods are often fitted for imaging problems, and were not tested on different kinds of data and constraints.

Despite the existence of many related prior works with different focuses, our goal is different and more challenging: obtaining samples distributed according to the target conditional distribution, while previous methods rather aim at obtaining high quality samples. To our knowledge, there are no works focusing on the correct approximation of the conditional distribution for tabular data as we do. The best comparison we can do is with \citet{bansal2023universal}, since it is the state of the art for diffusion based zero-shot conditional generation and a synthesis of previous techniques. We show that our method is significantly better with tabular data when the objective is the approximation of the conditional distribution.

\section{Method}

\subsection{Problem formalization}
Given a set of observed samples $\bvector{x}_i \in \mathbb{R}^d$, the goal is to sample from the distribution $p(\bvector{x})$ that generated $\bvector{x}_i$ conditioning on a desired property. Let $\pi(\bvector{x}): \mathbb{R}^d \rightarrow \{0,1\}$ be the function that encodes this property, such that $\pi(\bvector{x})=1$ when the property is satisfied and $\pi(\bvector{x})=0$ otherwise. Then the target conditional distribution can be defined as:
$$ p(\bvector{x} | \pi) \propto p(\bvector{x})\pi(\bvector{x})$$
Alternatively, one can also define soft constraints, expressing the degree of satisfaction as a real number. Let $c(\bvector{x}): \mathbb{R}^d \rightarrow \mathbb{R}$ be a differentiable function expressing this soft constraint. 
In this case we define the target distribution as:
$$p^c(\bvector{x}) \propto p(\bvector{x})e^{c(\bvector{x})}$$
Moreover, since the form is analogous to the previous formulation, given a hard constraint $\pi(\bvector{x})$ one can build a soft constraint $c(\bvector{x})$ such that $p^c(\bvector{x}) \approx p(\bvector{x}|\pi)$.
We then consider $p^c(\bvector{x})$ as the target distribution we want to sample from.

\subsection{Constraint-based guidance}
Our method exploits score-based generative models as the base generative model. As previously introduced, a stochastic process that gradually adds noise to original data $q(\tilde{\bvector{x}} | \bvector{x})$ is defined such that at $t=0$ no noise is added so $X_0 \sim p(\bvector{x})$ and at $t=1$ the maximum amount of noise is added such that $X_1 \sim q_1(\tilde{\bvector{x}} | \bvector{x})$ is a known prior distribution (for example a Gaussian). Given the possibility to efficiently sample from $p_t(\bvector{x})$, the time-dependent (\textit{Stein}) score of $p_t(\bvector{x})$ is estimated by score matching using a neural network, let it be $\bvector{s}(\bvector{x}, t) \approx \nabla_{\bvector{x}}\ln{p_t(\bvector{x})}$. As discussed in the previous section, there are different possible sampling schemes once the score is available. Given the target distribution:
$$p^c(\bvector{x}) := \frac{p(\bvector{x})e^{c(\bvector{x})}}{Z}$$
where $Z$ is the unknown normalization constant, and the distribution of samples from $p^c(\bvector{x})$ that are successively corrupted by $q(\tilde{\bvector{x}} | \bvector{x})$:
$$p^c_t(\bvector{x}) := \int{q_t(\bvector{x}|\bvector{x}_0) p^c(\bvector{x}_0) d\bvector{x}_0}$$
we observe the following relationship:
$$\nabla_{\bvector{x}} \ln{p^c_0(\bvector{x})} = \nabla_{\bvector{x}} \ln{p^c(\bvector{x})} = \nabla_{\bvector{x}} \ln{\frac{p(\bvector{x})e^{c(\bvector{x})}}{Z}}$$
$$= \nabla_{\bvector{x}} [\ln{p(\bvector{x})} + c(\bvector{x}) - \ln{Z}] = \nabla_{\bvector{x}} \ln{p(\bvector{x})} + \nabla_{\bvector{x}} c(\bvector{x})$$
It follows that at $t=0$ one can easily obtain an estimate of the score by summing the gradient of the constraint to the estimate of the score of the unconstrained distribution. Notice that this is possible since taking the gradient of the logarithm eliminates the intractable integration constant $Z$. At $t=1$ instead one can assume $\nabla_{\bvector{x}} \ln{p^c_1(\bvector{x})} = \nabla_{\bvector{x}} \ln{p_1(\bvector{x})}$, since it is reasonable to assume enough noise is added to make samples from $p^c_1(\bvector{x})$ distributed as the prior. In general there is no analytical form for $\nabla_{\bvector{x}} \ln{p^c_t(\bvector{x})}$, also, it cannot be estimated by score matching since we are not assuming samples from $p^c_0(\bvector{x})$ are available in the first place.


\subsection{Conditional score approximation}
Given this limit, we resort to approximations $\tilde{\bvector{s}}_c(\bvector{x}, t)$ for $\bvector{s}_c(\bvector{x}, t) = \nabla_{\bvector{x}} \ln{p^c_t(\bvector{x})}$.
The approximations we use are constructed knowing the true value of the score for $t=0$ and $t=1$:
\begin{align}
    \tilde{\bvector{s}}_c(\bvector{x}, 0) &= \bvector{s}(\bvector{x}, 0) + \nabla_{\bvector{x}} c(\bvector{x}) \\
    \tilde{\bvector{s}}_c(\bvector{x}, 1) &= \bvector{s}(\bvector{x}, 1)
\end{align}
A simple way to obtain this is by weighting the contribution of the gradient of the constraint depending on time:
$$\tilde{\bvector{s}}_c(\bvector{x}, t) = \bvector{s}(\bvector{x}, t) + g(t) \nabla_{\bvector{x}} c(\bvector{x})$$
where $g(t): [0,1] \rightarrow [0,1]$ satisfies $g(0) = 1$ and $g(1) = 0$. This is equivalent to extending the domain of the constraint to noisy data points $c(\bvector{x},t)$ and then approximating it with  $c(\bvector{x},t)=g(t)c(\bvector{x})$.
Sampling from the target distribution then reduces to substituting the score of the base model with the modified score $\tilde{\bvector{s}}_c(\bvector{x}, t)$. 
Notice that this approach does not require any re-training of the model.
The only necessary ingredients are the unconditional score model $\bvector{s}(\bvector{x}, t)$ and the differentiable constraint $c(\bvector{x})$ encoding the degree of satisfaction of the desired property. 

\subsubsection{Multiple instances constraints}
We may also want to sample multiple instances $\bvector{v} = (\bvector{x}_1, \ldots, \bvector{x}_n)$ that are tied together by a single multivariate constraint $c(\bvector{v})$, i.e., sampling from:
$p^c(\bvector{v}) \propto 
p(\bvector{v}) e^{c(\bvector{v})}
= e^{c(\bvector{v})} \prod_{i=1}^n p(\bvector{x}_i)$.
In this case, it is easy to show that $\nabla_{\bvector{x}_i} \ln{p^c_0(\bvector{x}_i)} = \nabla_{\bvector{x}_i} \ln{p(\bvector{x}_i)} + \nabla_{\bvector{x}_i} c(\bvector{x}_1, \ldots, \bvector{x}_n)$, then the approximated score for each instance $\bvector{x}_i$, $i \in \{1, \ldots, n\}$ is:
$$\tilde{\bvector{s}}_c(\bvector{x}_i, t) = \bvector{s}(\bvector{x}_i, t) + g(t) \nabla_{\bvector{x}_i} c(\bvector{x}_1, \ldots, \bvector{x}_n)$$
This can be computed in parallel for each instance $\bvector{x}_i$, as the computation of $\bvector{s}(\bvector{x}_i, t)$ can be parallelized by batching the score network and $\nabla_{\bvector{x}_i} c(\bvector{x}_1, \ldots, \bvector{x}_n)$ is just a component of $\nabla_{\bvector{v}} c(\bvector{v})$.
When sampling, the instances will also be generated in parallel, as they were a single instance. 

\subsubsection{Langevin MCMC correction.}
Depending on the type of data, we found it useful to perform additional Langevin dynamics steps (these are referred to as ``corrector steps" in \citet{sde}) at time $t = 0$ when the score of the constrained target distribution is known without approximation.  Langevin dynamics can be used as a Monte Carlo method for sampling from a distribution when only the score is known \cite{parisi1981correlation}, performing the following update, where $\epsilon$ is the step size and $\bvector{z}^i$ is sampled from a standard normal with the dimensionality of $\bvector{x}$: 
$$ \bvector{x}^{i+1} = \bvector{x}^i + \epsilon \,\tilde{\bvector{s}}_c(\bvector{x}, 0) + \bvector{z}^i \sqrt{2 \epsilon}$$
 In the limit $i \rightarrow \infty$ and $\epsilon \rightarrow 0$ Langevin dynamics samples from $p^c(\bvector{x})$. Nevertheless, as most Monte Carlo techniques, Langevin dynamics struggles in exploring all the modes when the target distribution is highly multimodal. 
Algorithm \ref{alg:constraint_guidance} summarizes the modified sampling algorithm. 

\subsubsection{Choice of score approximation scheme.}
We observed the results to be sensitive to the choice of the approximation scheme for $\nabla_{\bvector{x}} \ln{p^c_t(\bvector{x})}$. One should choose a $g(t)$ that is strong enough to guide samples towards the modes of $p^c_0(\bvector{x})$ but at the same time that does not disrupt the reverse diffusion process in the early steps.
We experimented with various forms of $g(t)$, mostly with the following two functions:
\begin{itemize}
    \item \textbf{Linear}: $g(t) = 1 - t$
    \item \textbf{SNR}: $g(t)$ is equal to the signal-to-noise ratio of the diffusion kernel. For example, if the diffusion kernel $q_t(\tilde{\bvector{x}} | \bvector{x})$ is $\mathcal{N}(\bvector{x}, \sigma_t)$, then $g(t) = (1 + \sigma_t^2)^{-\frac{1}{2}}$, assuming normalized data.
\end{itemize}
In many of our experiments we found $\textbf{SNR}$ to be the most effective, so we suggest using it as the first choice.

\begin{algorithm}[tbh]
\caption{Constraint guidance sampling}\label{alg:cap}
\label{alg:constraint_guidance}
\begin{algorithmic}
\State \textbf{Input:} constraint $c(\bvector{x})$, score $\bvector{s}(\bvector{x},t)$, score-based sampling algorithm $A(\bvector{s})$ 
\State \textbf{Parameters:} $g(t)$, $\epsilon$, $n$
\smallskip
\State $\tilde{\bvector{s}}_c(\bvector{x}, t) \gets \bvector{s}(\bvector{x},t) + g(t)\nabla_\bvector{x} c(\bvector{x})$
\State $\bvector{x} \gets A(\tilde{\bvector{s}}_c)$

\For{$i = 1$ \textbf{to} $n$}
    \State $\bvector{z} \gets \mathcal{N}(0,1)$ (with the dimensionality of $\bvector{x}$)
    \State $\bvector{x} \gets \bvector{x} + \epsilon  \tilde{\bvector{s}}_c(\bvector{x}, 0) + \bvector{z} \sqrt{2\epsilon}$
\EndFor

\State \textbf{return} $\bvector{x}$

\end{algorithmic}
\end{algorithm}

\subsection{Neuro-Symbolic logical constraints}
We will consider a general class of constraints expressed in a logical form. Hard logical constraints cannot be directly used in the approach presented above, hence we turn them into differentiable soft constraints leveraging neuro-symbolic ideas \cite{LTN}. More specifically, we consider predicates and formulae defined on the individual features $\bvector{x}=(x_1,\ldots,x_d)$ of the data points we ought to generate. Given a Boolean property $P(\bvector{x})$ (i.e. a predicate or a formula), with features $\bvector{x}$ as free variables, we associate with it a constraint function $c(\bvector{x})$ such that $e^{c(\bvector{x})}$ approximates the corresponding non-differentiable hard constraint $\mathbf{1}_{P(\bvector{x})}$.\footnote{$\mathbf{1}_{P(\bvector{x})}$ is indicator function equal to 1 for each $\bvector{x}$ such that $P(\bvector{x})$ is true.} 
In this paper, we consider constraints that can be evaluated on a single or on a few data points, hence we can restrict ourselves to the quantifier-free fragment of first-order logic. Therefore, we can define by structural recursion the constraint $c(\bvector{x})$ for atomic propositions and Boolean connectives. As atomic propositions, we consider here simple equalities and inequalities of the form $a(\bvector{x}) \geq b(\bvector{x})$, $a(\bvector{x}) \leq b(\bvector{x})$, $a(\bvector{x}) = b(\bvector{x})$, where $a$ and $b$ can be arbitrary differentiable functions of feature variables $\bvector{x}$. 
Following \citet{LTN}, we refer to such sets of functions as \textit{real logic}.

In particular,  we define the semantics directly in log-probability space, obtaining the \textbf{Log-probabilistic logic}.  The definitions we adopt are partially in line with those of the newly defined \textit{LogLTN}, see~\citet{Badreddine2023logLTNDF}, and are reported in Table~\ref{tab:logic}.

\begin{table}[ht]
\begin{tabular}{ll}
\textbf{Formula} & \textbf{Differentiable function} \\
\hline
$c[a(\bvector{x}) \geq b(\bvector{x})]$    &        $-\ln(1 + e^{-k(a(\bvector{x})-b(\bvector{x}))})$           \\
$c[a(\bvector{x}) \leq b(\bvector{x})]$    &        $-\ln(1 + e^{-k(b(\bvector{x})-a(\bvector{x}))})$          \\
$c[a(\bvector{x}) = b(\bvector{x})]$       &        $a(\bvector{x}) \geq b(\bvector{x}) \land a(\bvector{x}) \leq b(\bvector{x})$ \\
\hline
$c[\varphi_1 \land \varphi_2]$ & $c[\varphi_1] + c[\varphi_2]$ \\
$c[\varphi_1 \lor \varphi_2]$           & $\ln(e^{c[\varphi_1]} + e^{c[\varphi_2]} - e^{c[\varphi_1] + c[\varphi_2]})$  \\
$c[\lnot \varphi]$            & $\ln(1 - e^{c[\varphi]})$
\end{tabular}
\caption{Semantic rules of log-probabilistic logic. In the table, $c[\varphi](\bvector{x})$ is the soft constraint associated with the formula $\varphi$.}
\label{tab:logic}
\end{table}

\subsubsection{Atomic predicates.}
We choose to define the inequality $a(\bvector{x}) \geq b(\bvector{x})$ as $c(\bvector{x}) = -\ln(1 + e^{-k(a(\bvector{x})-b(\bvector{x}))})$, introducing an extra parameter $k$ that regulates the "hardness" of the constraint. Indeed, in the limit $k \rightarrow \infty$ one have $\lim_{{k \to \infty}} e^{c(\bvector{x})} = \mathbf{1}_{a(\bvector{x}) \geq b(\bvector{x})}$.  This definition is consistent with the negation but its gradient is nonzero when the condition is satisfied,\footnote{ This can be addressed by defining a simplified version: $a(\bvector{x}) \geq b(\bvector{x}) \equiv k(a(\bvector{x})-b(\bvector{x})) \textbf{1}_{a(\bvector{x}) <b(\bvector{x})}$, however such a definition will no more be consistent with negation.} though this was not creating issues in the experiments for sufficiently large values of $k$.
For the equality we use the standard definition based on inequalities.
We also experimented with a definition based on the l2 distance, corresponding to a Gaussian kernel, even if this form does not benefit from a limited gradient.


\subsubsection{Boolean connectives.}
The conjunction and the disjunction correspond to the product t-norm and its dual t-conorm (probabilistic sum) \cite{van2022analyzing} but in logarithmic space.
We use the material implication rule to reduce the logical implication to a disjunction: $a \rightarrow b \equiv \lnot a \lor b$.
The negation, instead, is consistent with the semantic definition of inequalities: negating one inequality, one obtains its flipped version.
For numerical stability reasons, however, we choose to avoid using the soft negation function in any case. Instead, we reduce logical formulas to the negation normal form (NNF) as in \citet{Badreddine2023logLTNDF}, where negation is only applied to atoms, for which the negation can be computed analytically or imposed by definition.
In order to simplify the notation, in the following we will also use quantifiers ($\forall$ and $\exists$) as syntactic sugar (in place of \textit{finite} conjunctions or disjunctions) only when the quantified variable takes values in a finite and known domain (e.g. time instants in a time series or pixels in an image).
So $\forall i \in \{1,\ldots, n\} \ p_i$ is used as a shorthand for $p_1 \land p_2 \land ... \land p_n$ and $\exists i \in \{1,\ldots, n\}: p_i$ for $p_1 \lor p_2 \lor ... \lor p_n$.

The difference in our definition with respect to \textit{LogLTN} is in the logical disjunction ($\lor$): they define it using the LogMeanExp (LME) operator, an approximation of the maximum that is numerically stable and suitable for derivation.
They do it at the price of losing the possibility to reduce formulas to the NNF exactly (using De Morgan's laws) that follows from having as disjunction the dual t-conorm of the conjunction. We choose instead to use the log-probabilistic sum as the disjunction, since in the domain of our experiments it proved numerically stable and effective. In Appendix \ref{app:disjunction} we show how to implement the logical disjunction in a numerically stable way. 

\smallskip
When sampling, we can regulate the tradeoff between similarity with the original distribution and strength of the constraint by tuning the parameter $k$ of inequalities in log-probabilistic logic.
Alternatively, we can multiply by a constant $\lambda$ the value of the constraint in order to scale its gradient.


\subsection{Training a score model for tabular data}
Score-based models have been mainly used for image generation, adapting them to tabular data and time series requires special care. In particular, the challenge is to correctly model the noise of the target distribution, implying the tricky task of estimating the score at $t \approx 0$ (no noise). We improved the score estimate mainly by parametrizing carefully the score network and using large batches. We provide more details in Appendix \ref{app:training}.
Correctly estimating the score at $t\approx0$ is fundamental to make our method work in practice, since it allows us to perform Langevin MCMC at $t\approx0$, where the conditional score is known without approximation. 
Combining a correct score estimation at $t\approx0$ with many steps of Langevin MCMC allows us to (asymptotically) sample from the exact conditional distribution, in particular when the data distribution is not particularly multimodal.

\section{Experiments}
We tested our method on several datasets, still, evaluating the quality of conditionally generated samples is challenging. First of all, one should compare conditionally generated samples with another method to generate conditionally in an exact way. We chose then to compare our approach with rejection sampling (RS), that can be used to sample exactly from the product of two distributions $p(\bvector{x})$ and $q(\bvector{x})$, where sampling from $p(\bvector{x})$ is tractable and $q(\bvector{x})$ density is known up to a normalization constant. In our case $p(\bvector{x})$ is the unconditional generative model and $q(\bvector{x}) = e^{c(\bvector{x})}$. Assuming constraints are such that $\forall{\bvector{x}}\, c(\bvector{x}) \leq 0$, then $q(\bvector{x})$ is upper-bounded by $1$. This upper bound is guaranteed by the real logic we defined previously. RS then reduces to sampling from $p(\bvector{x})$ and accepting each sample with probability $q(\bvector{x})$. This can be problematic when the probability of a random sample from $p(\bvector{x})$ having a non-null value of $q(\bvector{x})$ is low.
In the second place, comparing the similarity of two samples is a notoriously difficult problem. 
For relatively low dimensional samples, we will compare the marginal distributions and the correlation matrix.
For comparing one-dimensional distributions among two samples $X$ and $Y$ we use the l1 histogram distance
$ D(X,Y):=\frac{1}{2} \sum_i |x_i - y_i|$ where $x_i$ and $y_i$ are the empirical probabilities for a given common binning. This distance is upper bounded by 1.
When computationally feasible, we consider RS as the baseline method. By reporting the acceptance rate of RS, we show the satisfaction rate of the constraint on data generated by the original model.
Moreover, we discuss in Section \ref{sec:ug_comparison} a comparison with \citet{bansal2023universal} that is arguably the state of the art method for zero-shot conditional generation of images.

We mostly used unconditional models based on denoising score matching and SDEs, following closely \citet{sde}. Conditional generation hyperparameters and more details about models are described in Appendix \ref{app:settings} and \ref{app:unconditional_models}.


\subsection{Tabular data}\label{sec:white_wine}

We made experiments with the white wine table of the popular UCI Wine Quality dataset \cite{misc_wine_quality_186}, consisting of 11 real-valued dimensions ($\mathbb{R}^{11}$) and one discrete dimension, the quality, that we discarded.
In order to evaluate the effectiveness with categorical variables, we also made experiments with the Adult dataset \cite{misc_adult_2}, consisting of 5 numerical dimensions and 10 categorical dimensions, which we embedded in a continuous space using one-hot encodings. We fitted unconditional score-based diffusion models based on SDEs, then we generated samples under illustrative logical constraints.

\begin{figure}[ht]
\centering
  \includegraphics[width=0.95\columnwidth]{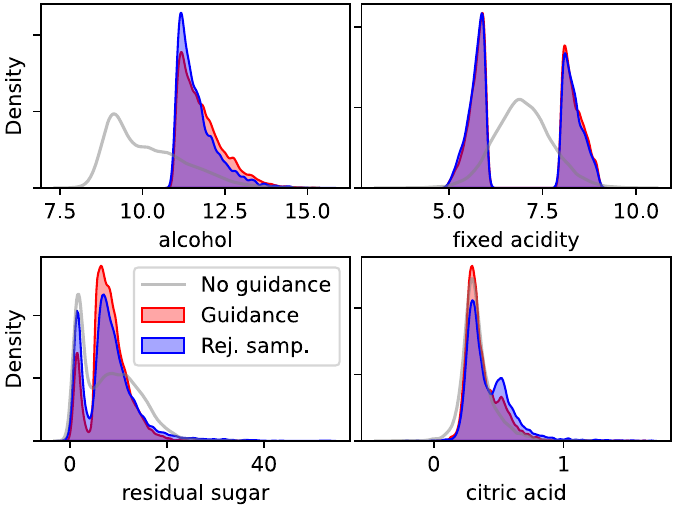}
  \caption{Marginals of white wine data experiment. We generated 5000 samples using our constrained sampling algorithm and as many by RS. The plot compares the marginals of the dimensions directly involved in the constraint. The last two dimensions are the ones with the largest l1 histogram distance with respect to RS marginals: $0.15$ and $0.13$. While the median distance across all dimensions is $\approx 0.1$. In order to evaluate the noise of the distance, we also measured the self-distance between two equally sized samples obtained by RS and observed a median across dimensions of $\approx 0.05$. 
  }
\label{fig:white_wine}
\end{figure}

First, we generated samples from the white wine model under the following complex logical constraint: $(\text{fixed acidity} \in [5.0, 6.0]  \lor  \text{fixed acidity} \in [8.0, 9.0]) \land \text{alcohol} \geq 11.0 \land (\text{residual sugar} \leq 5.0 \rightarrow \text{citric acid} \geq 0.5$).
We show in \figref{fig:white_wine} the marginals of the generated samples, compared with 
samples generated by RS. 
There is a high overlap between marginals for most dimensions, and we measured an average l1 distance between correlation coefficients of $\approx 0.07$. The largest error, that is associated with one of the dimensions heavily affected by the constraint, is still relatively small. For that constraint, the acceptance rate of RS was only $\approx  1.67\%$, meaning that our  method samples efficiently in low-probability regions.
The satisfaction rates of the relative hard constraint were similar: $\approx 92\%$ for RS and $\approx 86\%$ for our method (these can be increased by increasing the parameter $k$).
Additionally, we tested the application of a simple multi-instance constraint acting on pair of data points $\bvector{x}_1$ and $\bvector{x}_2$: $\text{alcohol}_1 > \text{alcohol}_2 + 1$. Comparing again with RS, we achieved a negligible error and the relative hard constraint was met in $99\%$ of the generated samples, compared to the $100\%$ of RS, with an acceptance rate of $27\%$.
We further confirmed the effectiveness of our method by generating samples from the Adult model under the following logical constraint: $\text{age} \geq 40 \land (\text{race} \neq \text{"White"} \lor \text{education} = \text{"Masters"})$. In this case equalities and inequalities involving discrete components are obtained by imposing the desired component of the corresponding one-hot encoding equal to 1 for equality, or to 0 for inequality. The median l1 histogram distance was $\approx 0.05$, the maximum was $\approx 0.13$ and the error in correlations was negligible. The relative hard constraint was met in all samples while the acceptance rate of RS was $1.73\%$.





\subsection{Time series surrogate models} \label{sec:eSIRS} 
A surrogate model is a simplified and efficient representation of a complex, eventually computationally expensive model. It is possible to learn a surrogate model of a complex stochastic dynamical system by fitting a statistical model to a dataset of trajectories observed from it. 
Following our approach, one can use a score-based generative model to learn an unconditional surrogate model, and then apply constraints to enforce desirable properties. These can be physical constraints the system is known to respect, or features that are rare in unconditioned samples. So we can exploit this method to both assure consistency of trajectories and explore rare (but not necessarily with low density) scenarios.
As a case study, we apply our proposed method for the conditional generation of ergodic SIRS (eSIRS) trajectories. The eSIRS model \cite{kermack1927} is widely used to model the spreading of a disease in an open population\footnote{Open in the sense of having infective contacts with external individuals, not part of the modelled population.}. The model assumes a fixed population of size $N$ composed of Susceptible ($S$), Infected ($I$), and Recovered ($R$) individuals. We consider trajectories with $H$ discretized time steps, thus we have that the sample space is $\mathcal{X}_{\textbf{eSIRS}} := {\big({\mathbb{N}_0}^2\big)}^{H}$, where the two dimensions are $S$ and $I$ ($R$ is implicit since $R = N - S - I$).

First we train a score-based generative model to fit trajectories that were generated by a simulator, with 
$H=30$ and $N=100$.
Then we experimented with the application of different constraints, including the following consistency constraints:
\begin{itemize}
    \item \textit{Non-negative populations}: $\forall t \ S(t) \geq 0 \land I(t) \geq 0$

    \item \textit{Constant population}: $\forall t \ S(t)+I(t) \leq N$
\end{itemize}



\begin{figure}[ht]
\centering
  \includegraphics[width=0.85\columnwidth]{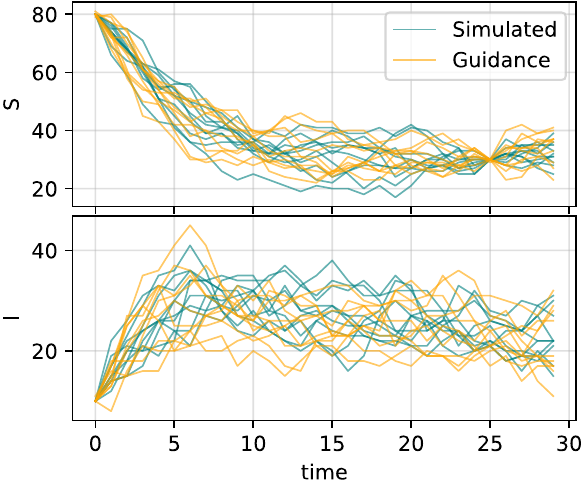}
  \caption{Bridging with eSIRS time series. We show here a subsample of the $5000$ time series generated with constraint guidance (orange) with a subsample of the $>50000$ time series generated by RS from the simulator (green). Additionally to the consistency constraints, that are always met, we imposed the following equalities: $S(0)=95$, $I(0)=5$, $S(25)=30$. Constraints are generally met: the average l1 absolute difference with all three target values is below $0.19$. The l1 histogram distance for each time step marginal is relatively small, considering that it accounts also for the error of the unconditional model: for $S$ and $I$ the median l1 histogram distance across time are $\approx 0.11$ and $\approx 0.13$. The median self distance across time between two samples of 5000 instances of RS was $\approx 0.04$ for both $S$ and $I$.}
  \label{fig:eSIRS_bridging}
\end{figure}

\begin{figure}[ht]
\centering
  \includegraphics[width=0.85\columnwidth]{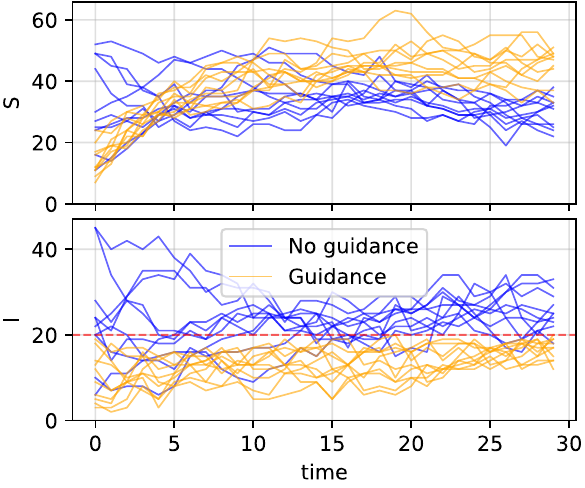}
  \caption{Imposing an inequality on eSIRS time series. We show here a subsample of the 100 time series generated with and without constraint guidance. Additionally to the consistency constraints, that are always met, we imposed $\forall t \ I(t) \leq 20$, that is perfectly met in $99\%$ of the samples.}
  \label{fig:eSIRS_inequality}
\end{figure}

We show in \figref{fig:eSIRS_bridging} and \figref{fig:eSIRS_inequality} two experiments with two different constraints. In both experiments the consistency constraints (positive and constant population) were always met, with a small improvement over the unconditional model. In the two experiments we additionally imposed a bridging constraint and an inequality, that were also met with minimal error.

\subsection{Images}
We test our method on image datasets in order to investigate the potential in high-dimensional data. %
We consider satisfactory validating 
the results by visual inspection of the generated images, since in this case the quality of individual samples is often considered more important than matching the true underlying distribution.
We do not use classifier-based  metrics such as FID or Inception score
since data samples from the original conditioned distribution are not available, hence a comparison is not possible.

We use as pre-trained unconditional models a model based on a U-net that we trained on the  MNIST dataset, and a pre-trained model for CelebA \cite{liu2015deep} 64x64 images made available in \citet{song2020improved}.

\paragraph{Digits sum.}
Using a pre-trained MNIST classifier, we define a multi-instance constraint that forces pairs of mnist digits to sum up to ten. Given pairs of images $(\bvector{x},\bvector{y})$, we define the constraint in the following way:
    $$\bigvee_{i=1}^{9} \text{{class}}(\bvector{x}, i) \land \text{{class}}(\bvector{y}, 10-i) $$
    where $\text{class}(\bvector{x},i):= P\{\bvector{x} \text{ is classified as } i\} = 1$, and $P$ 
    is obtained from a pre-trained classifier.
The generated pairs of digits add up to ten in $\approx 96\%$ of cases\footnote{according to classes assigned by the classifier}, however, only 2-8 and 4-6 pairs were generated. 
More details are discussed in Appendix \ref{app:other_experiments}.

\paragraph{Restoration.} Given a differentiable  function $f(\cdot)$, that represents a corruption process in which information is lost, we define the following constraint:
$$\forall i \ f(\bvector{x})_i = \tilde{\bvector{y}}_i$$
where $i$ is the pixel index and $\tilde{\bvector{y}}$ is a corrupted sample, possibly such that there is a $\bvector{y}$ that satisfies $\forall i \ \tilde{\bvector{y}}_i \approx f(\bvector{y})_i$. Such constraint has the effect of sampling possible $\bvector{x}$ such that $\forall i \ f(\bvector{x})_i = \tilde{\bvector{y}}_i$, i.e., ``inverting" $f$ or reconstructing the original $\bvector{y}$. $f$ can be any degradation process, such as downsampling, blurring or adding noise. So our approach can be used for image restoration for a known degradation process.

\begin{figure}[ht]
    \centering
  \includegraphics[width=0.9\columnwidth]{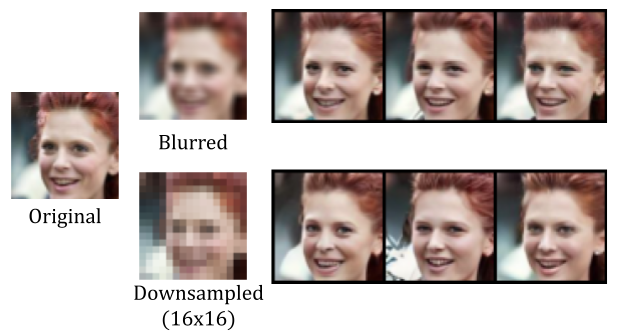}
  \caption{Restoration experiments with CelebA. The first image on the left is a sample image from the CelebA dataset. Each row shows the corrupted image followed by samples generated imposing the restoration constraint.}
  \label{fig:celebA_restoration}
\end{figure}

In \figref{fig:celebA_restoration} we show the results of image restoration experiments with a blurred and a downsampled image. Samples are realistic and we report a low error with respect to the target corrupted image (the absolute error per channel is approximately less than $0.015$).

In general, we observed the effectiveness of our method to depend on the task. For most experiments involving tabular data, keeping a large constraint strength $k$ (as $k=30$) was sufficient. However, for some experiments, as most tasks involving images, we needed to tune the constraint strength until a satisfactory trade-off between constraint satisfaction and quality of samples was reached.
In Appendix \ref{app:other_experiments} we discuss more experiments and cases where this trade-off was poorer.

\subsection{Comparison with universal guidance}
\label{sec:ug_comparison}
We compared our 
conditioning method with Universal Guidance, introduced in \citet{bansal2023universal}. Universal Guidance was successfully used in the context of image generation and it is arguably the state of the art for zero-shot conditional generation. Their method is based on three improvements over the standard guidance technique, 
that they call \textit{forward universal guidance}, \textit{backward universal guidance} and \textit{per-step self-recurrence}. \textit{Forward universal guidance} consists in using the gradient of the constraint  with respect to the predicted clean data point $\bvector{\hat{x}}_0$ (prediction based on the trained denoising net), instead of the current data point $\bvector{x}_t$. \textit{Backward universal guidance} involves an optimization of $\bvector{\hat{x}}_0$ according to the constraint and modifying the score used for the reverse process as a consequence.
Applying these techniques on tabular data and time series we observed little effect of \textit{forward universal guidance}, and a significantly detrimental effect of \textit{backward universal guidance}.  
Metrics reported in Table \ref{tab:universal_guidance_comparison_short} show the superiority of our method for two of the previously discussed tasks. We think this significant difference is due to the fact that Universal Guidance, as most recent plug-and-play techniques, modify the sampling algorithm  
by introducing an optimization phase.

\begin{table}[ht]
\begin{tabular}{@{}lcccc@{}}
\toprule
\multicolumn{1}{c}{\multirow{2}{*}{Method}} & \multicolumn{2}{c}{White wine} & \multicolumn{2}{c}{eSIRS bridging} \\ \cmidrule(l){2-5} 
\multicolumn{1}{c}{} & \textbf{Avg L1} & \textbf{Max L1} & \textbf{Avg L1} & \textbf{Max L1} \\ \midrule
\textbf{Ours}        & \textbf{0.1}    & \textbf{0.15}  & \textbf{0.13}  & \textbf{0.25}  \\
Univ. Guidance                   & 0.29           & 0.85           & 0.47           & 1.0             \\ \bottomrule
\end{tabular}
\caption{Comparison with Universal Guidance. On the first white wine and eSIRS bridging tasks our method performs significantly better in terms of l1 histogram distance (we report average and maximum distance over the marginals).}\label{tab:universal_guidance_comparison_short}
\end{table}

On the other side, 
we noticed a sensible improvement in some of our image-based conditional generation tasks. For example, relatively to the MNIST digits' sum experiment, universal guidance allowed us to generate all possible pairs, that always summed up to ten. 
We infer that universal guidance can be useful when the objective is to generate high-quality samples that satisfy a given constraint, but it can lead to large biases when modelling the conditional distribution, that is of primary interest in some contexts, as with tabular data and time series.
Moreover, with these experiments we verified that our \textit{Log-probabilistic logic} 
is effective also within other guidance techniques. We discuss the experiments more deeply in Appendix \ref{app:universal_guidance}.

\section{Conclusion}

We have shown how we can exploit pre-trained unconditional score-based generative models to sample under user-defined logical constraints, without the need for additional training. Our experiments demonstrate the effectiveness in several contexts, such as tabular and high-dimensional data. Nevertheless, in some high dimensional settings as images, we had to trade-off between sample quality and constraint satisfaction by tuning the constraint strength. More sophisticated methods specifically designed for images are probably necessary, still these could fail to model the true conditional distribution.
In fact, we show the superiority of our method in approximating conditional distributions for tabular and time-series data with respect to a state of the art method for zero-shot conditioning. 
Future work will aim at finding better approximation schemes, starting from works focused on general plug-and-play guidance for images. 

\section*{Acknowledgements}
This research was supported by Aindo, which has funded the PhD of the first author and provided the computational resources. Additionally, this study was carried out within the PNRR research activities of the consortium iNEST funded by the European Union Next-GenerationEU (PNRR, Missione 4 Componente 2, Investimento 1.5 – D.D. 1058 23/06/2022, ECS\_00000043).


\bibliography{aaai25}

\iftrue
\onecolumn

\appendix

\section*{Appendix}

\renewcommand{\thesubsection}{\Alph{subsection}}

\section{Comparison with Logic Tensor Networks}\label{app:LTN}

Deep learning allows to learn complex statistical properties at the cost of data inefficiency, struggle at generalization, lack of comprehensibility and ignorance of prior knowledge. In contrast, Symbolic AI has proven useful at several reasoning tasks (as theorem proving, logical inference, verification) but at the cost of high computational complexity and struggle with large and inaccurate data. Neurosymbolic AI aims at combining the strength of both paradigms.
Recent research has focused on developing differentiable approaches for knowledge representation and reasoning. This includes relaxing logical operators into continuous operations using fuzzy semantics.
Logic Tensor Networks \cite{LTN} is an example of such a neurosymbolyc framework, that supports learning and reasoning about data with a knowledge base. In LTN, one can represent and perform several tasks of deep learning (as clustering, classification, embedding learning) with a fully differentiable first order logic language, that they call Real Logic, assuming truth values in the interval [0,1].
Real Logic employs fuzzy semantics, where logical connectives ($\land$, $\lor$, ...) and quantifiers ($\forall$, $\exists$) are mapped to fuzzy operators like t-norms and aggregators. These operators enable formulas to have degrees of truth in the interval [0, 1], which are optimized during training by maximizing the satisfaction of a knowledge base of logical rules.
In LTN the preferred fuzzy logic is the Stable Product Real Logic.
LogLTN \cite{Badreddine2023logLTNDF} is a recent extension aiming to improve numerical stability in fuzzy semantics by working in logarithm space, as maximizing the log satisfaction of a formula is equivalent to maximizing its satisfaction.

\smallskip

Our LogProbabilistic logic has many similarities with the fuzzy logic used in LogLTN. We devolped it independently mainly motivated by probabilistic principles, as our formulas can be seen as describing un-normalized distribution encoding prior knowledge. In practice, the difference with their approach lies in the mapping of the logical disjunction. While we used the probabilistic disjunction as shown in table \ref{tab:logic}, they used a surrogate of the maximum function, the LogMeanExp operator, as the maximum function is single-passing (the gradient on passed through one of the arguments):

$$\operatorname{LME}(\mathbf{x} \mid \alpha, C)=\frac{1}{\alpha}\left(C+\log \left(\frac{\sum_{i=1}^n e^{\alpha x_i-C}}{n}\right)\right)$$
where $C = \max_i(\alpha x_i)$ and $\alpha$ is an hyperparameter. Authors claim that the LME operator is numerically stable, well-bounded, and suitable for derivation.
Our numerically stable implementation of the logical disjunction that we discuss in Appendix \ref{app:disjunction} performed well in our experiments,  being at the same time well grounded in probabilistic principles.

\section{Unconditional models}\label{app:unconditional_models}

For all datasets except CelebA we trained unconditional generative models from scratch. We used score based generative models based on SDEs \cite{sde}, also following many of the implementation details of the repository of the original paper.
We used two types of SDEs in our experiments: the variance exploding SDE (VE) and a variant of the variance preserving SDE, the sub-VP SDE. As score-matching algorithm we mostly used denoising score-matching and experimented also with sliced score matching. As numerical solver of the reverse SDE we used Euler-Maruyama, the simplest method. For most of our experiments with SDE we used predictor-corrector sampling, i.e, adding steps of Langevin dynamics after each step of reverse SDE solving. As discussed previously we often performed additional Langevin MCMC steps at $t=0$.
For the experiments with celebA we used a pre-trained model made available in the repository relative to \citet{song2020improved}, that is based on annealed Langevin dynamics.
For tabular data and time series the neural network architectures we used were simple multi layer perceptrons.
Regarding the model trained on the Adult dataset, we dealt with categorical variables by embedding them in continuous space using one-hot encodings. At generation time, one-hot encodings were transformed back to categorical variables by taking the argmax. Similarly, we generated integer variables by rounding to the closest integer. 
More complete information can be found in the code repository that we made available at \href{https://github.com/DavideScassola/score-based-constrained-generation}{\texttt{https://github.com/DavideScassola/score-based-constrained-generation}}.

\section{Training a score model for tabular data}\label{app:training}

In \citet{song2020improved} they suggest scaling the neural network approximating the noise dependent score in the following way: $\bvector{s}_\theta(\bvector{x}, t) = \frac{\bvector{f}_\theta(\bvector{x})}{\sigma_t}$ where $\bvector{f}_\theta(\bvector{x})$ is the neural network and $\sigma_t$ is the standard deviation of the diffusion kernel $q(\tilde{\bvector{x}} | \bvector{x})$. The issue is that when $t \approx 0$, $\sigma_t$ will approach zero and the score will explode. Apparently this is not an issue for image datasets, where modelling the exact distribution is less important than the quality of samples. This is not the case for time series and tabular data, where we are interested in correctly modelling the noisy distribution that generated the data. Moreover this is particularly important when we perform Langevin MCMC at $t\approx0$ in order to exploit the knowledge of the exact conditional score. We then elaborated some practices in order to improve the estimate in these domains:
\begin{itemize}
    \item We capped the scaling factor $\sigma_t^{-1}$ by a certain limited value. This value has to be an hyperparameter of the training.
    \item We used large batch sizes and occasionally accumulated the gradient across several batches before performing the weights update. This helps reducing the noise of the score-matching loss for low values of $\sigma_t$.
    \item We occasionally used sliced score matching to learn the noise dependent score on data corrupted by the diffusion process. The reason is that we found the sliced score matching loss to be less noisy than the denoising score matching loss, especially at $t \approx 0$.
\end{itemize}

\section{Normalization}\label{app:normalization}

Data in experiments was normalized. Since the constraint is defined in the original data space, one has to take the normalization into account when applying the constraint to transformed data. Given a data point $\bvector{x}$, a differentiable and invertible transformation $\bvector{f}(\bvector{x})$ and the constraint $c_x(\bvector{x})$, the actual constraint that is applied to transformed data points $\bvector{y} = \bvector{f}(\bvector{x})$ is $c_y(\bvector{y}) = c_x(\bvector{f}^{-1}(\bvector{y}))$.

\section{Constrained generation settings}\label{app:settings}
In Table \ref{tab:hyperparameters} are shown hyperparameters that are used in the constrained generation experiments. In experiments that involved equality constraints, we specify which function we used: LPL (log-probabilistic logic) refers to the first definition we gave in Table \ref{tab:logic}, otherwise we used the squared error. We also indicate the values of $k$ and $\lambda$ we used to regulate the constraints' intensity. The step size for Langevin MCMC was chosen dynamically depending on the norm of the score, we used the same implementation as in \citet{sde}. Constraints' intensity was chosen by hand, monitoring the trade-off between constraint satisfaction and samples quality across multiple experiments. The score approximation strategy ($g(t)$) was less influential than the constraint intensity, although we generally observed 
better performances when using SNR.

\section{Numerically stable implementation of logical disjunction}\label{app:disjunction}
Given the values of two soft constraints $x, y \in (-\infty, 0]$ according to the Log-probabilistic logic defined above, we defined the logical disjunction as: $$\lor(x,y) := \ln(e^x + e^y - e^{x + y})$$
From the practical point of view, when both $x$ and $y$ are small the argument of the logarithm can became $0$, thus causing a numerical issue.
We tackle this by rewriting the formula in the following analytically equivalent way:

$$\ln(1 + e^{y-x} - e^{y}) + x \ \ \text{when} \ x>y $$

$$\ln(e^{x-y} + 1 - e^{x}) + y \ \ \text{when} \ y>x $$

More concisely:

$$\lor(x,y) := \ln(1 + e^{\text{min}(x,y) - \text{max}(x,y)} - e^{\text{min}(x,y)}) + \text{max}(x,y) $$

\section{Universal guidance comparison experiments}\label{app:universal_guidance}

Applying the three universal guidance techniques \cite{bansal2023universal} in some of the experiments described above, we observed the following:
\begin{itemize}
    \item On tabular data and time series, the effect of forward universal guidance was mostly negligible, except for models trained with sliced score matching, where it led to failure. Thus, for our experiments with universal guidance we used only models trained with denoising score matching.
    \item Backward universal guidance and per-step self-recurrence instead led to a significant worsening of the performances in tabular and time series data. We show in Table \ref{tab:wine_universal_guidance_comparison} and Table \ref{tab:bridging_backward_guidance} statistics about the comparison. \figref{fig:white_wine_backward_guidance} shows how using backward guidance introduces strong biases in some marginals in the white wine experiment. We think this may be due to the optimization process involved, that focuses on optimizing samples but ignores the noise, thus failing to the converge to the true conditional distribution. We also verified the effect of Langevin MCMC correction in these experiments, and we  observed a clear reduction of the bias introduced by the application of backward universal guidance, compared with the same configuration without Langevin correction steps.

    \begin{figure}[ht]
    \centering
      \includegraphics[width=0.5\columnwidth]{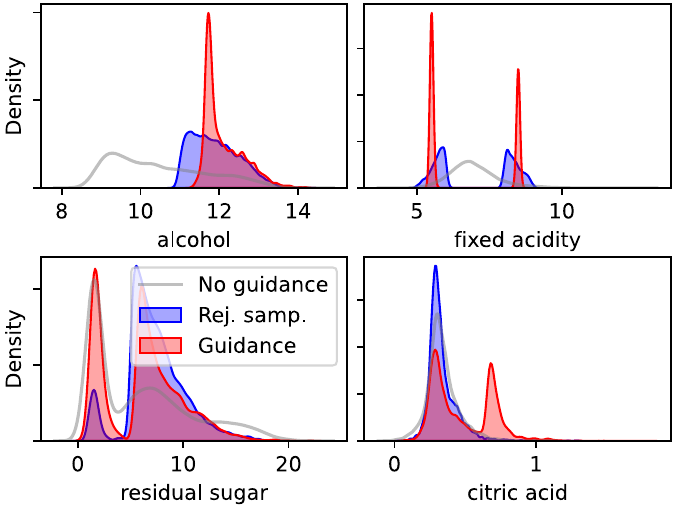}
      \caption{Effect of \textit{backward universal guidance} on white wine experiment. We generated 5000 samples using forward universal guidance and backward universal guidance (without Langevin MCMC correction) and as many by RS. The plot compares the marginals of the dimensions directly involved in the constraint. We can see that the distribution of generated samples is highly biased.} 
      \label{fig:white_wine_backward_guidance}
    \end{figure}

    \begin{table}[ht]
    \centering
    \begin{tabular}{lcccc}
    \toprule
    \textbf{Method}     & \textbf{Average L1 HD} & \textbf{Median L1 HD} & \textbf{Max L1 HD} & \textbf{Average L1 corr.} \\ \midrule
    SSM + Langevin MCMC                & 0.1                    & 0.1                   & 0.153              & 0.07                              \\
    DSM                                & 0.1                    & 0.081                 & 0.205              & 0.059                             \\
    DSM + FW                           & 0.123                  & 0.112                 & 0.262              & 0.059                             \\
    DSM + FW + BW                      & 0.238                  & 0.13                  & 0.827              & 0.093                             \\
    DSM + FW + BW + SR                 & 0.288                  & 0.162                 & 0.850              & 0.094                             \\
    DSM + Langevin MCMC                & 0.132                  & 0.13                  & 0.221              & 0.084                             \\
    DSM + FW + Langevin MCMC           & 0.127                  & 0.114                 & 0.232              & 0.082                             \\
    DSM + FW + BW + Langevin MCMC      & 0.141                  & 0.117                 & 0.3                & 0.088                             \\
    DSM + FW + BW + SR + Langevin MCMC & 0.167                  & 0.124                 & 0.377              & 0.099                             \\ \bottomrule
    \end{tabular}
    
        \caption{Applying universal guidance to white wine experiment. We compare here the application of universal guidance techniques (FW: forward, BW: backward and SR: per-step self-recurrence) to our method for the white wine constrained generation task discussed in Section \ref{sec:white_wine}. The results previously discussed above were obtained with a model trained with sliced score matching (SSM), but in order to apply universal guidance, we also trained a model with denoising score matching (DSM) and reduced the strenght of the constraint (from $k=50$ to $k=30$). We found forward universal guidance to have little effect on the results. Instead, using backward universal guidance significantly increased the error in terms of histogram distance (HD) and also in terms of average l1 distance of correlation coefficients. Average, mean and maximum are computed aggregating the errors of the marginals of all the 11 dimensions.}\label{tab:wine_universal_guidance_comparison}
    \end{table}


    \begin{table}[ht]
    \centering

    \begin{tabular}{lccc}
    \toprule
    \textbf{Method}                    & \textbf{Average L1 HD} & \textbf{Median L1 HD} & \textbf{Max L1 HD} \\ \midrule
    DSM                                & 0.152                  & 0.142                 & 0.371              \\
    DSM + FW                           & 0.113                  & 0.102                 & 0.221              \\
    DSM + FW + BW                      & 0.197                  & 0.153                 & 1.0                \\
    DSM + FW + BW + SR                 & 0.467                  & 0.485                 & 1.0                \\
    DSM + Langevin MCMC                & 0.129                  & 0.114                 & 0.252              \\
    DSM + FW + Langevin MCMC           & 0.107                  & 0.094                 & 0.208              \\
    DSM + FW + BW + Langevin MCMC      & 0.126                  & 0.116                 & 0.288              \\
    DSM + FW + BW + SR + Langevin MCMC & 0.399                  & 0.342                 & 0.77               \\ \bottomrule
    \end{tabular}
    \caption{Applying universal guidance to eSIRS bridging experiment. We compare here the application of universal guidance techniques to our method for the eSIRS bridging task discussed in \figref{fig:eSIRS_bridging}. We found forward universal guidance to have little but positive effect on the results, instead using backward universal guidance increased the error, especially when Langevin MCMC correction was not applied. Average, mean and maximum are computed aggregating the errors on the marginals of all dimensions ($S(t), I(t)$).}
    \label{tab:bridging_backward_guidance}
    \end{table}
     
    \item In MNIST experiments, the application of forward guidance led to an improvement in the digits' sum task, as we observed all digits between 1 and 9 in the generated samples (instead of only 2-8, 4-6 couples generated with our method) and a higher quality (see \figref{fig:digits_sum_experiment_with_universal_guidance} and \figref{fig:mnist_digits_sum}). On image symmetry experiments instead the generated images were mostly ones ($\approx 87.5\%$ with $\approx 15.6\%$ of invalid digits). 

    \begin{figure}[ht]
    \centering
      \includegraphics[width=0.6\columnwidth]{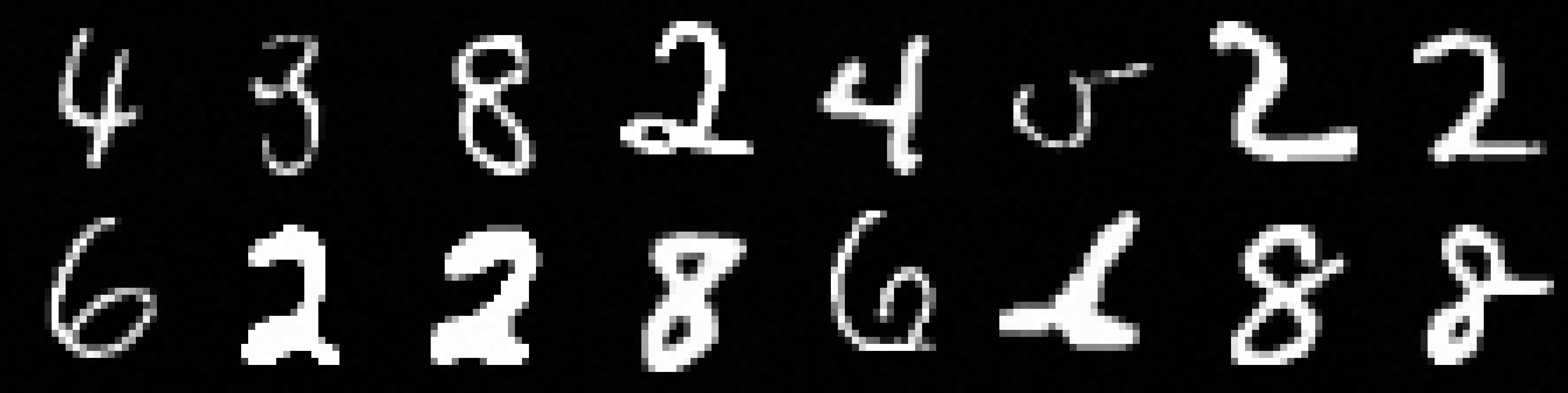}
      \caption{Digits sum experiment. The columns are the pairs of generated digits. Using our method we were only able to generate 2-8 and 6-4 pairs. The classes assigned by the classifier added up to ten in $96\%$ of cases, but visual inspection reveals $\approx 12\%$ of invalid digits.}
      \label{fig:mnist_digits_sum}
    \end{figure}

    \begin{figure}[ht]
    \centering
      \includegraphics[width=0.6\columnwidth]{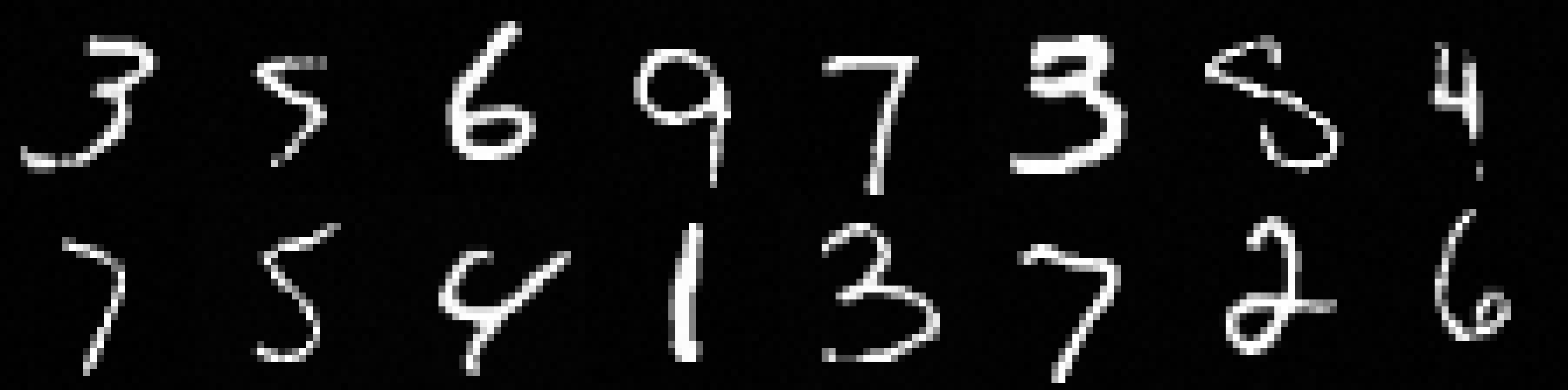}
      \caption{Digits sum experiment with universal guidance. The columns are the pairs of generated digits. Using forward universal guidance we were able to generate all possible pairs that summed up to ten. The classes assigned by the classifier added up to ten in $100\%$ of cases, but visual inspection reveals $\approx 8\%$ of invalid digits. Using also backward universal guidance and per-step self-recurrence led to very similar results.}
      \label{fig:digits_sum_experiment_with_universal_guidance}
    \end{figure}

\end{itemize}



\section{Other experiments}\label{app:other_experiments}

\paragraph{White wine: multiple instances constraint.}
We defined a constraint involving pairs of instances of white wine data $(\bvector{x}_1, \bvector{x}_2) $: 
$\text{alcohol}(\bvector{x}_1) > \text{alcohol}(\bvector{x}_2)+1$. Our method was able to generate samples that follow closely the correct target distribution, as we can see from the low errors in terms of l1 histogram distance of the marginals and average l1 distance of correlation coefficients reported in Table \ref{tab:white_wine_multi}. The relative hard constraint was met in $99\%$ of the generated samples, compared to the $27\%$ of the unconditioned model.

\begin{table}[ht]
\centering
\begin{tabular}{lcccc}
\toprule
$\text{alcohol}(\textbf{x}_1) > \text{alcohol}(\textbf{x}_2)+1$ & \textbf{Average L1 HD} & \textbf{Median L1 HD} & \textbf{Max L1 HD} & \textbf{Average L1 corr.} \\ \midrule
$\bvector{x}_1$                                  & 0.06                   & 0.058                 & 0.074              & 0.025                             \\
$\bvector{x}_2$                                  & 0.061                  & 0.063                 & 0.086              & 0.026                             \\ \bottomrule
\end{tabular}
\caption{Multiple instances constraint on wine dataset. Considering the constraint $\text{alcohol}(\bvector{x}_1) > \text{alcohol}(\bvector{x}_2)+1$, the first row shows the errors measure relative to the dimensions of the first element of the pair $\bvector{x}_1$, while the second shows them for $\bvector{x}_2$.}
\label{tab:white_wine_multi}
\end{table}



\paragraph{MNIST: Symmetry.}
We impose vertical or horizontal symmetry constraints to generated images, by imposing pixel-by-pixel equality between the image and its mirrored version. In \figref{fig:mnist_symmetry} we show resulting samples using a pre-trained model for MNIST images.

\begin{figure}[ht]
\centering
  \includegraphics[width=0.50\columnwidth]{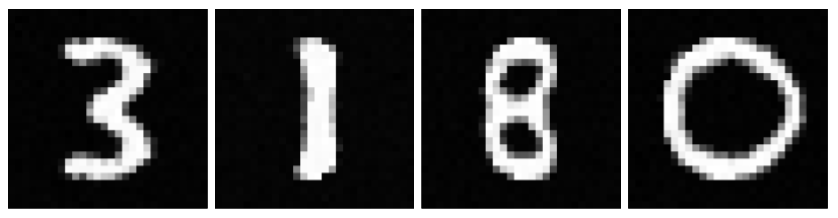}
  \caption{MNIST samples generated imposing a horizontal symmetry constraint. In many cases ($\approx 30\%$) the generated image did not correspond to an actual digit, this is probably due to the unconditional model, that also generated a significant amount of invalid digits ($\approx 10\%$).}
  \label{fig:mnist_symmetry}
\end{figure}

\paragraph{MNIST: relative filling.} We construct constraints that force the generated image to have more white pixels on a certain part of the image with respect to the remaining. In order to do this, we impose the average pixel value of a given region to be larger than the average pixel value of the remaining pixels by a value of $1.0$ (considering normalized images). We show in \figref{fig:mnist_relative_filling} some samples generated imposing this constraint, in particular imposing the average pixel value of the upper half to be greater of the average of the lower part by one (or vice-versa). 
\begin{figure}[ht]
\centering
  \includegraphics[width=0.5\columnwidth]{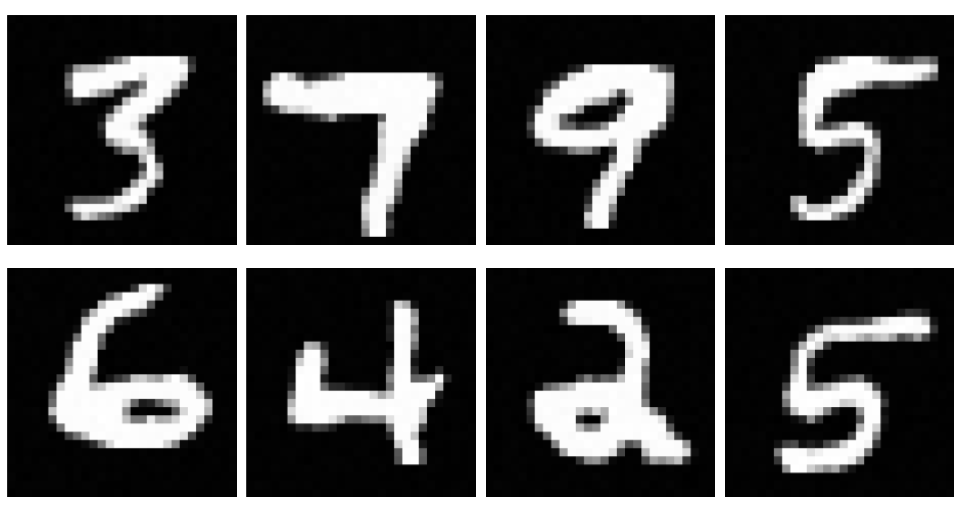}
  \caption{Relative filling experiments. The upper row of digits is generated imposing more white pixels on the upper half than the lower, the lower row is generated imposing more white pixels on the lower half.}
  \label{fig:mnist_relative_filling}
\end{figure}

\paragraph{CelebA: color conditioning.} With the objective of controlling the coloring of samples, we use a simple equality constraint between a target RGB color and the mean RGB color of the image. One can also think about more sophisticated constraints, that aim at matching a target distribution in the color space. The resulting samples shown in \figref{fig:celebA_color_conditioning} demonstrate the possibility to
manipulate the coloring of an image while preserving the
quality of the samples. Nevertheless, it seems the color is
excessively manipulated through background, hair, shades
or a “filter” effect. This is probably due to the naive approach
of targeting the mean color.

\begin{figure}[ht]
\centering
  \includegraphics[width=0.5\columnwidth]{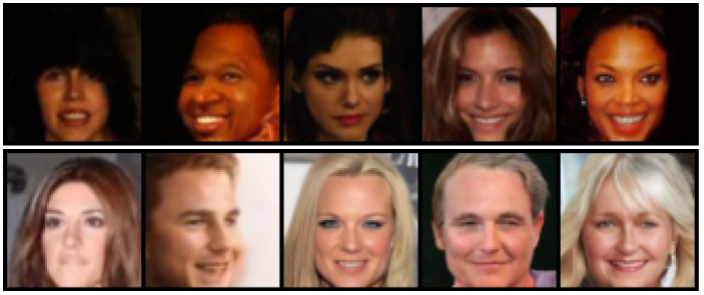}
  \caption{Color conditioning on CelebA. The first row of images is generated imposing the mean color to be equal to a given dark target color, extracted from an image of the dataset. For the second row instead the target was a light color.}
  \label{fig:celebA_color_conditioning}
\end{figure}

\paragraph{CelebA: symmetry.} The constraint we used in order to obtain images with vertical symmetry is analogous to the one we used for MNIST digits. \figref{fig:celebA_symmetry} shows how the symmetry soft constraint is successfully met in most samples, and how perfect symmetry is traded-off with realism. As in most tasks involving images, we found necessary to tune the constraints' intensity $k$ or $\lambda$, in order to regulate the trade-off between sample quality and constraint satisfaction.

\begin{figure}[ht]
\centering
  \includegraphics[width=0.5\columnwidth]{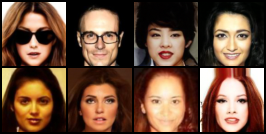}
  \caption{CelebA samples generated imposing a vertical
symmetry constraint. Notice the bias for frontal face position and uniform background. While the constraint was useful to improve the symmetry of images, reaching a perfect symmetry by increasing the constraint intensity was not possible without compromising the samples' quality.} 
  \label{fig:celebA_symmetry}
\end{figure}

\paragraph{Example on a toy dataset.} We fitted a score-based generative model on data generated according to a mixture of two Gaussian distributions $\mathcal{N}(\mu=-3,\,\sigma=0.5)$, $ \mathcal{N}(\mu=4,\,\sigma=1)$ with equal mixing coefficients.
Given the constraint $x \geq 0$, our method generates samples that are not distinguishable from ones generated with RS. We show in \figref{fig:toy} the distributions generated without guidance, with constraint guidance and with RS.

\begin{figure}[ht]
\centering
  \includegraphics[width=0.5\columnwidth]{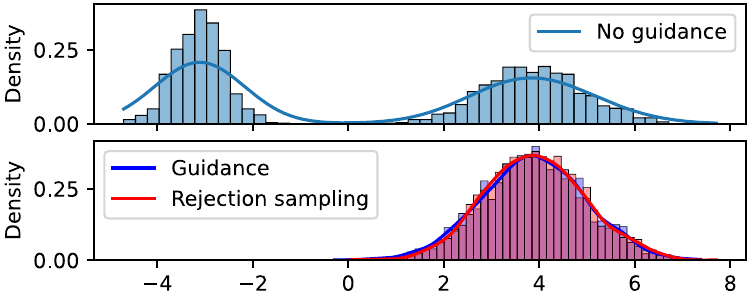}
  \caption{Constraint guidance on a toy dataset.} We fitted a score-based generative model on data generated according to a mixture of two Gaussian distributions $\mathcal{N}(\mu=-3,\,\sigma=0.5)$, $ \mathcal{N}(\mu=4,\,\sigma=1)$ with equal mixing coefficients.
  \label{fig:toy}
\end{figure}

\begin{table}[ht]
\centering
\begin{tabular}{l|r}
\hline
\textbf{Column} & \textbf{l1 histogram distance} \\
\hline
Fixed Acidity & 0.049 \\
Volatile Acidity & 0.052 \\
Citric Acid & 0.13 \\
Residual Sugar & 0.15 \\
Chlorides & 0.077 \\
Free Sulfur Dioxide & 0.094 \\
Total Sulfur Dioxide & 0.11 \\
Density & 0.10 \\
pH & 0.087 \\
Sulphates & 0.12 \\
Alcohol & 0.11 \\
\hline
\end{tabular}
\caption{White wine experiment with a complex constraint: l1 histogram distance for different columns. This table compares the marginal distributions of the 5000 samples generated with constraint guidance with the same number of samples generated with RS. The median across-dimensions of the measured self distance between two distinct samples of 5000 instances generated by RS was of $\approx 0.05$   }
\label{tab:marginals_distance}
\end{table}

\section{Hardware specifications}
\begin{itemize}
    \item Processor: AMD Ryzen Threadripper 2950X 16-Core Processor
    \item Memory: 125 GiB
    \item GPU: NVIDIA RTX A5000
\end{itemize}

\begin{table}
\centering
\begin{tabular}{@{}llll@{}}
\toprule
\textbf{Experiment} & \textbf{Constraint} & \textbf{g(t)} & \textbf{Langevin MCMC steps} \\ 
\midrule
White wine: complex constraint (SSM) & LPL($k=50$) & Linear & 4000 \\
\midrule
White wine: complex constraint (DSM) & LPL($k=30$) & SNR & 2000 \\
\midrule
Adult: logical formula (DSM) & LPL($k=30$) & SNR & 10 \\
\midrule
White wine: multivariate constraint (SSM) & LPL($k=50$) & SNR & 4000 \\
\midrule
eSIRS bridging & \begin{tabular}[c]{@{}l@{}}LPL,\\ equality: $k = 7$\\ consistency: $k = 1$\end{tabular} & SNR & 2000 \\
\midrule
eSIRS inequality & \begin{tabular}[c]{@{}l@{}}LPL,\\ inequality: $k = 25$\\ consistency: $k = 1$\end{tabular} & SNR & 2500 \\
\midrule
MNIST digits sum & LPL($k=30$) & SNR & 10 \\
\midrule
MNIST relative filling & LPL($k=80$) & SNR & 20 \\
\midrule
MNIST symmetry & LPL($k=5$) & SNR & 0 \\
\midrule
CelebA symmetry & Squared error ($\lambda=0.19$) & SNR & 0 \\
\midrule
CelebA color conditioning & \begin{tabular}[c]{@{}l@{}}Squared error,\\ light: $\lambda = 1000$\\ dark: $\lambda = 800$\end{tabular} & SNR & 0 \\
\midrule
CelebA restoration & \begin{tabular}[c]{@{}l@{}}Squared error,\\ deblurring: $\lambda = 100$\\ upsampling: $\lambda = 600$\end{tabular} & SNR & 0 \\
\midrule
Toy example & LPL($k=50$) & SNR & 200 \\
\bottomrule
\end{tabular}
\caption{Constrained generation details}
\label{tab:hyperparameters}
\end{table}

\fi
\end{document}